\documentclass{article}
\usepackage{spconf, amsmath, graphicx}
\usepackage{amssymb}
\usepackage{color}
\usepackage{booktabs}
\newcommand{\best}[1]{{\color{black}{\textbf{#1}}}} 

\newcommand{\vect}[1]{\boldsymbol{#1}}
\newcommand{\methodname}{{\tt{IAA-LQ}}}

\title{Image Aesthetics Assessment via Learnable Queries}
\name{Zhiwei Xiong$^{1,2}$, Yunfan Zhang$^{1,2}$, Zhiqi Shen$^1$, Peiran Ren$^3$, Han Yu$^1$
\thanks{This research is supported, in part, by the National Research Foundation, Singapore and DSO National Laboratories under the AI Singapore Programme (AISG Award No: AISG2-RP-2020-019); Alibaba Group through Alibaba Innovative Research (AIR) Program and Alibaba-NTU Singapore Joint Research Institute (JRI) (Alibaba-NTU-AIR2019B1), Nanyang Technological University, Singapore; and the RIE 2020 Advanced Manufacturing and Engineering (AME) Programmatic Fund (No. A20G8b0102), Singapore.}}
\address{\textit{$^1$School of Computer Science and Engineering, Nanyang Technological University, Singapore}\\
\textit{$^2$Alibaba-NTU Singapore Joint Research Institute, Nanyang Technological University, Singapore}\\
\textit{$^3$Alibaba Group, Hangzhou, China}}
\begin{document}
%
\maketitle
\begin{abstract}
Image aesthetics assessment (IAA) aims to estimate the aesthetics of images. Depending on the content of an image, diverse criteria need to be selected to assess its aesthetics. Existing works utilize pre-trained vision backbones based on content knowledge to learn image aesthetics. However, training those backbones is time-consuming and suffers from attention dispersion. Inspired by learnable queries in vision-language alignment, we propose the \underline{I}mage \underline{A}esthetics \underline{A}ssessment via \underline{L}earnable \underline{Q}ueries (\methodname{}) approach. It adapts learnable queries to extract aesthetic features from pre-trained image features obtained from a frozen image encoder. Extensive experiments on real-world data demonstrate the advantages of \methodname{}, beating the best state-of-the-art method by 2.2\% and 2.1\% in terms of SRCC and PLCC, respectively.
\end{abstract}
\begin{keywords}
Aesthetics Assessment, Learnable Queries
\end{keywords}
\section{Introduction}
\label{sec:intro}

Image aesthetics assessment (IAA) is a computer vision task that aims to evaluate the aesthetic quality of images. Such a capability is beneficial for downstream applications including image recommendation, enhancement, retrieval, and generation~\cite{zhang2021comprehensive}. Due to the inherent subjectivity and ambiguity associated with image aesthetics, the ground truth of image aesthetics is usually determined by the opinions of different reviewers in the form of the mean opinion score (MOS) or the distribution of opinion scores (DOS).

Depending on the content of the image, there can be different emphases involved in aesthetics assessment. Early works proposed to split the images into different semantic groups and extract different sets of aesthetic features~\cite{tang2013content, kao2016hierarchical, wang2016multi}. The features can be handcrafted under the guidance of photography rules~\cite{tang2013content}, or extracted by deep neural networks~\cite{kao2016hierarchical, wang2016multi}. However, there may be special cases where an image does not belong to any predefined semantic group. Explicitly splitting these images based on semantics can result in the relationships among different semantic groups being overlooked.

Therefore, later works~\cite{hosu2019effective, he2022rethinking, xiong2023federated} attempted to implicitly extract aesthetic features from pre-trained semantic backbones. Since fine-tuning the entire backbone is computationally expensive~\cite{hosu2019effective, xiong2023federated} and might lead to attention dispersion~\cite{he2022rethinking}, the backbone is only used as a feature extractor. However, since the vision backbones are pre-trained primarily for image classification~\cite{hosu2019effective, xiong2023federated} or scene recognition~\cite{he2022rethinking}, they lack knowledge regarding the aesthetic attributes (e.g., composition) that are less related to semantics. Moreover, some of these works~\cite{hosu2019effective, xiong2023federated} require input images in full resolution and additional feature extraction stages in advance, which lack efficiency and practical applicability.

More recently, with the prevalence of pre-trained large vision-language models, works that utilize such models together with specially designed prompts for IAA are starting to emerge~\cite{wang2023exploring, ke2023vila}. The vision-language models used are either pre-trained solely on general image-text pairs~\cite{wang2023exploring}, or further pre-trained on aesthetic image-text pairs~\cite{ke2023vila}. Either way, their pre-trained models are not limited to aesthetic-related semantic patterns but also cover relatively abstract knowledge related to image aesthetics. However, they use relatively simple single prompts (e.g., ``good image") or prompt pairs (e.g., ``good photo" and ``bad photo") to extract aesthetic-related knowledge, which cannot deal with complex IAA tasks.

Inspired by BLIP-2~\cite{li2023blip}, a vision-language pre-training model that uses learnable queries to align vision and language pre-trained features extracted from frozen unimodal models, we propose the \underline{I}mage \underline{A}esthetics \underline{A}ssessment via \underline{L}earnable \underline{Q}ueries (\methodname{}) approach.
It trains learnable queries to extract aesthetic features from the pre-trained vision backbones. With a flexible quantity, the learnable queries can extract the most aesthetics-related image patterns from the frozen vision backbone. Extensive experiments on real-world data demonstrate the advantages of \methodname{}, beating the best state-of-the-art method by 2.2\% and 2.1\% in terms of Spearman's rank correlation coefficient (SRCC) and Pearson linear correlation coefficient (PLCC), respectively.

\section{The Proposed \methodname{} Approach}
\label{sec:iaa-lq}
\begin{figure}[!b]
\centering
\includegraphics[width=0.95\linewidth]{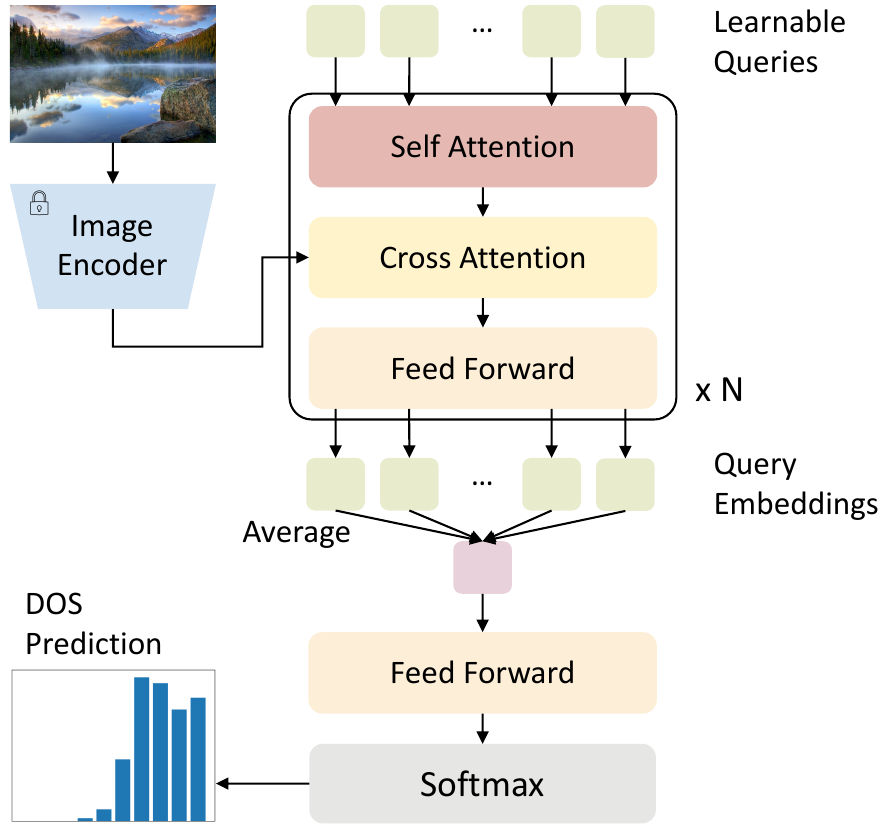}
\caption{The design of \methodname{}. It learns embeddings for learnable queries through a querying transformer, where pre-trained image features extracted with a frozen image encoder are inserted once in every two transformer blocks for cross-attention. The learned query embeddings are averaged and passed through a feed-forward layer and Softmax to output the predicted aesthetic DOS.} \vspace{-2.5mm}
\label{fig:iaa_lq}
\end{figure}

The proposed \methodname{} approach is shown in Figure~\ref{fig:iaa_lq}. It consists of a frozen image encoder to extract pre-trained image features, a set of learnable queries, a querying transformer for self- and cross-attention, and a prediction header for IAA.

\subsection{Encoding an Image}
A given image can be expressed as $(I_n, \vect{d}_n)$, where $I_n$ is the $n$-th image in an IAA dataset, and $\vect{d}_n = \{d_n^k\}_{k=1}^K$ is its ground truth aesthetic DOS which satisfies $\sum_{k=1}^K d_n^k = 1$. Note the aesthetic MOS $s_n$ can be derived from $\vect{d}_n$ as:
\begin{equation}
s_n = \sum_{k=1}^K (k \cdot d_n^k).
\end{equation}
In our design, we adopt pre-trained vision transformers (ViT) as the image encoder. It splits the input image into $N_p$ patch tokens and adds a \texttt{[CLS]} token as the first token. Therefore, the extracted pre-trained image embeddings $\vect{e}_v^n$ of image $I_n$ can be expressed as:
\begin{equation}
\vect{e}_v^n = E_{\theta_v}(I_n),
\end{equation}
where $E_{\theta_v}(\cdot)$ is the ViT parameterized by frozen $\theta_v$, and $\vect{e}_v^n \in \mathbb{R}^{(1+N_p) \times H_v}$ with $H_v$ denoting the hidden size of the ViT.

\subsection{Learnable Queries \& Querying Transformer}
Suppose we use $M$ learnable queries with hidden size $H_q$ to extract the aesthetic features from the pre-trained image features. The pre-trained image embeddings are inserted once every two transformer blocks into the querying transformer for cross-attention with the queries. In each block, the queries first interact with each other through self-attention, and then possibly conduct cross-attention with the image embeddings. Finally, they are passed through a feed-forward layer to output query embeddings. Formally, for queries $\vect{q} \in \mathbb{R}^{M \times H_q}$ and pre-trained image embeddings $\vect{e}_v^n$, the output query embeddings $\vect{e}_q^n$ with aesthetic knowledge extracted from the image is expressed as:
\begin{equation}
\vect{e}_q^n = E_{\theta_q}(\vect{q} \vert \vect{e}_v^n),
\end{equation}
where $E_{\theta_q}(\cdot)$ is the querying transformer parameterized by $\theta_q$, and $\vect{e}_q^n \in \mathbb{R}^{M \times H_q}$.

\subsection{Prediction Header for IAA}
After the output query embeddings $\vect{e}_q^n$ are obtained, we take the average of the query embeddings to obtain a compact aesthetic embedding:
\begin{equation}
\vect{e}_a^n = \Bar{\vect{e}}_q^n,
\end{equation}
where $\vect{e}_a^n \in \mathbb{R}^{H_q}$. Based on the aesthetic embedding, a prediction header for IAA is appended. It is a feed-forward layer with an input size of $H_q$ and an output size of $K$ followed by a Softmax layer. Together, the predicted $K$-scale aesthetic DOS can be computed as:
\begin{equation}
\hat{\vect{d}}_n = E_{\theta_p}(\vect{e}_a^n).
\end{equation}
$E_{\theta_p}(\cdot)$ is the prediction header parameterized by $\theta_p$. Following~\cite{ talebi2018nima}, we adopt Earth Mover’s Distance (EMD) loss to optimize the predicted DOS towards the ground truth:
\begin{equation}
\mathcal{L}(\vect{d}_n, \hat{\vect{d}}_n) = \sqrt{\frac{1}{K}\sum_{k=1}^K \lvert CDF_{\vect{d}_n}(k)-CDF_{\hat{\vect{d}}_n}(k) \rvert ^2}.
\end{equation}
$CDF_{\vect{d}_n}$ and $CDF_{\hat{\vect{d}}_n}$ are cumulative density functions for the ground truth DOS and predicted DOS, respectively. The overall objective is to minimize the loss over the whole dataset:
\begin{equation}
\min_{\vect{q}, \theta_q, \theta_p} \left\{\sum_{n=1}^N \mathcal{L}(\vect{d}_n, \hat{\vect{d}}_n) \right\}.
\end{equation}

\section{Experimental Evaluation}
\label{sec:experiments}

\subsection{Experiment Settings}

The dataset adopted for our experimental evaluation is the benchmark IAA dataset, the AVA dataset~\cite{murray2012ava}. It contains over 250,000 images, each of which received 78 to 549 aesthetic scores (average of 210) on a scale of 1 to 10 (i.e., $K=10$),
where 1 and 10 denote the lowest and highest aesthetics, respectively. 
The ground truth DOS could be derived by computing the ratio of reviewers under each score level to the total number of reviewers. 
The same train-test split in~\cite{hosu2019effective, ke2023vila} is adopted for our experiments, where 235,574 and 19,928 images are allocated for training and testing, respectively. We also include the generic testing set of the PARA dataset~\cite{yang2022personalized} to evaluate the generalization ability and attribute-level performance of \methodname{}. It contains 3,000 images with overall and attribute-level aesthetic annotations.

\begin{table}[!b]
\centering
\small
\begin{tabular}{lcc}
\toprule
Method &SRCC &PLCC \\
\midrule
NIMA \cite{talebi2018nima}                      &0.612 &0.636 \\
AFDC + SPP \cite{chen2020adaptive}              &0.649 &0.671 \\
GPF-CNN \cite{zhang2019gated}                   &0.671 &0.682 \\
MaxViT \cite{tu2022maxvit}                      &0.708 &0.745 \\
MUSIQ \cite{ke2021musiq}                        &0.726 &0.738 \\
MLSP \cite{hosu2019effective}                   &0.756 &0.757 \\
TANet \cite{he2022rethinking}                   &0.758 &0.765 \\
GAT$_{\times3}$-GATP \cite{ghosal2022image}     &0.762 &0.764 \\
VILA-R \cite{ke2023vila}                        &0.774 &0.774 \\
\midrule
\methodname{}                                          &\best{0.791} &\best{0.790} \\
\bottomrule
\end{tabular}
\vspace{-1mm}
\caption{Experiment results on the AVA dataset.}
\label{tab:peer_compare}
\vspace{-3mm}
\end{table}

Following~\cite{li2023blip}, the pre-trained image embeddings are retrieved from the second last layer of the pre-trained ViT. The image resolution is set to $224 \times 224$ with a patch size of $14 \times 14$, yielding 257 image embeddings (i.e., $N_p = 256$). We set the query hidden size $H_q$ to 768, initialize the Q-Former with BERT$_\text{base}$~\cite{devlin2018bert}, and randomly initialize the cross-attention layers. We train the Q-Former with a batch size of 128 for 10 epochs with Adam optimizer. The learning rate is set to $3 \times 10^{-5}$ initially, and multiplied by $0.1$ every two epochs. To evaluate the performance of the comparison approaches, we report Spearman's rank correlation coefficient (SRCC) and Pearson linear correlation coefficient (PLCC) between the predicted and ground truth aesthetic MOSs.

\subsection{Comparison Results}

Table~\ref{tab:peer_compare} shows the performances of \methodname{} compared with 9 state-of-the-art (SOTA) methods under the AVA dataset. It can be observed that \methodname{} achieves the best performance, exceeding the best-performing SOTA method VILA-R~\cite{ke2023vila} by $2.2\%$ and $2.1\%$ in terms of SRCC and PLCC, respectively. To obtain this model, we employ Horizontal Flipping (HF) with $p=0.5$ on training images, use ViT-G/14 from EVA-CLIP~\cite{fang2023eva} as the frozen vision backbone which outputs pre-trained image embeddings with $H_v = 1408$, and $M = 2$ learnable queries to extract the aesthetic features. Similar to VILA-R, \methodname{} takes input images with a low resolution rather than the full resolution as required by previous works~\cite{hosu2019effective, ke2021musiq}. It demonstrates that \methodname{} can effectively learn from the most critical aesthetics-related image clues.

\subsection{Ablation Studies}

\noindent\textbf{Effects of image padding and augmentations}:
Since image aesthetics is empirically found to be sensitive to image paddings and augmentations, we investigate their effects in \methodname{}. The two most common augmentations used in IAA, Horizontal Flipping (HF) and Random Cropping (RC), are considered. As shown in Table~\ref{tab:augmentation}, RC deteriorates our model performance, which could be due to the content and composition changes that are highly related to image aesthetics as described in photographic rules such as the Rule of Thirds. On the other hand, HF brings slight performance improvement because it generally does not affect image aesthetics, while bringing in more samples to train on.

\begin{table}[!t]
\centering
\small
\begin{tabular}{cccc}
\toprule
Padding &Augmentation & SRCC & PLCC \\
\midrule
True    &None      &0.782  &0.783 \\
True    &HF        &\textbf{0.784}  &\textbf{0.785} \\
True    &RC        &0.781  &0.781 \\
True    &HF + RC   &0.780  &0.780 \\
\midrule
False   &None      &0.790  &0.790 \\
False   &HF        &\textbf{0.791}   &\textbf{0.790} \\
False   &RC        &0.784  &0.785 \\
False   &HF + RC   &0.785  &0.784 \\
\bottomrule
\end{tabular}
\vspace{-1mm}
\caption{Performance of \methodname{} with different padding and augmentation strategies. HF denotes Horizontal Flipping with $p=0.5$. RC denotes resizing to $272 \times 272$ followed by Random Cropping of $224 \times 224$.}
\label{tab:augmentation}
\end{table}


\begin{table}[!t]
\centering
\small
\begin{tabular}{c|c|cccc}
\toprule
M & 32 & 1 & 2 & 3 & 4 \\
\midrule
SRCC &0.768 &0.788 &\textbf{0.791} &0.790 &0.788 \\
PLCC &0.766 &0.788 &\textbf{0.790} &0.789 &0.788 \\
\bottomrule
\end{tabular}
\vspace{-1mm}
\caption{Performance of \methodname{} with different numbers of learnable queries. M denotes the number of queries.}
\label{tab:num_queries}
\end{table}

\begin{table}[!t]
\centering
\small
\begin{tabular}{cccc}
\toprule
Vision Backbone &Embeddings & SRCC & PLCC \\
\midrule
ViT$_\text{CLIP}$        &\texttt{CLS}        &0.735  &0.735 \\
ViT$_\text{CLIP}$        &\texttt{CLS+P}     &0.626  &0.628 \\
ViT$_\text{CLIP}$        &\texttt{LQ}          &\textbf{0.774}  &\textbf{0.775} \\
\midrule
ViT$_\text{EVA-CLIP}$    &\texttt{CLS}         &0.743  &0.743 \\
ViT$_\text{EVA-CLIP}$    &\texttt{CLS+P}    &0.713  &0.712 \\
ViT$_\text{EVA-CLIP}$    &\texttt{LQ}          &\textbf{0.791}  &\textbf{0.790} \\
\bottomrule
\end{tabular}
\vspace{-1mm}
\caption{Performance of \methodname{} with different vision backbones and different embeddings for the aesthetic prediction. \texttt{CLS} denotes the embedding of the \texttt{[CLS]} token. \texttt{CLS+P} denotes the average of the embedding of the \texttt{[CLS]} token and patch embeddings. \texttt{LQ} denotes the average of the learned query embeddings.}
\label{tab:backbone}
\vspace{-3mm}
\end{table}

\noindent\textbf{Effects of number of queries}:
The number of learnable queries $M$ is a crucial hyperparameter in \methodname{} as it directly affects the richness of the aesthetic features extracted. As shown in Table~\ref{tab:num_queries}, we start exploring the optimal $M$ by setting it to 32, the same as in BLIP-2~\cite{li2023blip}. However, it can be observed that with 32 queries, \methodname{} stops improving at very early stages and easily overfits. Therefore, we then attempt to set $M=1$ and increase it gradually. Experiment results demonstrate that \methodname{} achieves the best performance when only two queries are used. It demonstrates its ability of using a small number of queries to capture adequate aesthetic features for IAA.

\noindent\textbf{Effectiveness of the learnable queries}:
The frozen vision backbone determines the pre-trained image patterns that can be used for IAA. In our experiments, we consider ViT-L/14 from CLIP~\cite{radford2021learning} and ViT-G/14 from EVA-CLIP~\cite{fang2023eva}, which output image embeddings with $H_v = 1024$ and $H_v = 1408$, respectively. To demonstrate the effectiveness of the learnable queries, we compare the performances of learning image aesthetics from the pre-trained \texttt{[CLS]}  embedding, from the average of pre-trained \texttt{[CLS]} and patch embeddings, and from the query embeddings of the learnable queries. As shown in Table~\ref{tab:backbone}, the outstanding performance of the learnable queries demonstrates the effectiveness of the design of the \methodname{} learnable queries and querying transformer in extracting relevant aesthetic features from pre-trained image features.

\subsection{Model Interpretation}

\begin{figure}[!t]
\centering
\includegraphics[width=0.95\linewidth]{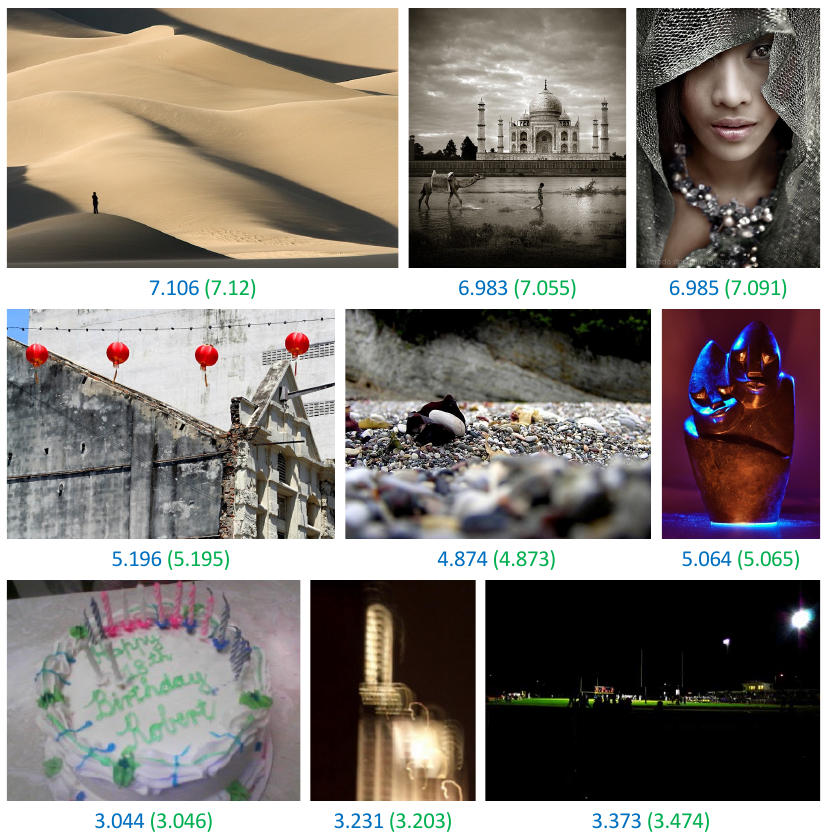}
\caption{Examples of the \methodname{} MOS prediction results. Images from the top row to the bottom row are example images with relatively high, moderate, and relatively low ground truth MOSs. The blue and (green) numbers beneath each image are its predicted and (ground truth) MOSs, respectively.} \vspace{-2.5mm}
\label{fig:pos_examples}
\end{figure}

In Figure~\ref{fig:pos_examples}, we show some examples of the aesthetic MOS prediction results using the proposed \methodname{} approach. 
The example images are retrieved from the testing set of AVA.
It can be observed that \methodname{} can estimate image aesthetics for various image contents, such as landscapes, objects, and portraits. However, it does suffer from increased prediction errors when the test image has relatively extreme aesthetic quality (either too high or too low). It can be attributed to the data distribution of the AVA dataset, where the majority of the images have aesthetic MOS around the borderline (i.e., 5). This can be improved by training \methodname{} on more images with relatively extreme aesthetics MOSs.

\begin{table}[t]
\centering
\small
\begin{tabular}{ccc}
\toprule
Attribute & SRCC & PLCC \\
\midrule
Aesthetics  &0.701  &0.739 \\
\midrule
Composition &0.702  &0.737 \\
Content     &0.686  &0.730 \\
Quality     &0.684  &0.725 \\
DOF         &0.667  &0.711 \\
Light       &0.642  &0.694 \\
Color       &0.619  &0.667 \\
\bottomrule
\end{tabular}
\vspace{-1mm}
\caption{Attribute-level performance of \methodname{}.}
\label{tab:attr}
\vspace{-3mm}
\end{table}

To evaluate the generalization capability of \methodname{} and its ability to indicate fine-grained aesthetic attributes, we directly perform inference on the testing set of PARA using \methodname{} and compare the estimated aesthetics MOSs with the ground truth MOSs of different aesthetic attributes. Table~\ref{tab:attr} shows that even without further supervision from the training set of PARA, \methodname{} can still estimate the overall aesthetic qualities reasonably well. The high correlation with the composition and content attributes could be credited to the relatively simple augmentations used and the frozen pre-trained image encoder. Interestingly, our predicted MOS has a relatively low correlation with the color attribute. This can be attributed to the large amount of near-gray low-saturation images in AVA. Nevertheless, the proposed \methodname{} approach demonstrates strong generalization capability and attribute-level aesthetics awareness.

\section{Conclusions and Future Work}
\label{sec:conclusions}


In this paper, we propose the \methodname{} approach, which learns image aesthetics using learnable queries. With a small number of learnable queries, our approach can effectively extract the most relevant aesthetic features from the pre-trained image features obtained with a frozen image encoder. Extensive experiments demonstrate that \methodname{} largely outperforms state-of-the-art approaches on the benchmark AVA dataset, and achieves strong generalization ability. 

In subsequent research, we plan to improve the explainability of \methodname{} with additional outputs in terms of aesthetic attributes and captions or additional logical reasoning in the model design.



\bibliographystyle{IEEEbib}
\bibliography{main}

\end{document}